\documentclass[accepted]{uai2024} % after acceptance, for a revised version; 
\usepackage[american]{babel}
\usepackage{xassoccnt}
\newcounter{realpage}
\DeclareAssociatedCounters{page}{realpage}
\AtBeginDocument{%
  \stepcounter{realpage}
}

\usepackage{natbib} % has a nice set of citation styles and commands
\usepackage{mathtools} % amsmath with fixes and additions
\usepackage{caption}
\usepackage{subcaption}
\usepackage{graphicx,enumerate,amssymb,bbold}
\usepackage{booktabs} % commands to create good-looking tables
\usepackage{tikz} % nice language for creating drawings and diagrams

\newcommand{\bitem}{\begin{itemize}}
\newcommand{\eitem}{\end{itemize}}
\newcommand{\mc}[1]{\mathcal{#1}}

\newcommand{\II}{\mathbb{I}}

\newcommand{\R}{\mathbb{R}}

\newcommand{\EE}{\mathbb{E}}

\newcommand{\PP}{\mathbb{P}}

\newcommand{\bpm}{\begin{pmatrix}}
\newcommand{\epm}{\end{pmatrix}}
\newcommand{\bvm}{\begin{vmatrix}}
\newcommand{\evm}{\end{vmatrix}}
\newcommand{\bsm}{\left(\begin{smallmatrix}}
\newcommand{\esm}{\end{smallmatrix}\right)}
\newcommand{\T}{\top}

\newcommand{\ol}[1]{\overline{#1}}
\newcommand{\wh}[1]{\widehat{#1}}
\newcommand{\wt}[1]{\widetilde{#1}}
\newcommand{\la}{\langle}
\newcommand{\ra}{\rangle}

\newcommand{\eins}{\mathbb{1}}

\DeclareMathSymbol{\mydiv}{\mathbin}{symbols}{"04}

\DeclareMathOperator{\Diag}{Diag}

\DeclareMathOperator{\Exp}{Exp}
\makeatletter
\def\widebreve{\mathpalette\wide@breve}
\def\wide@breve#1#2{\sbox\z@{$#1#2$}%
     \mathop{\vbox{\m@th\ialign{##\crcr
\kern0.08em\brevefill#1{0.8\wd\z@}\crcr\noalign{\nointerlineskip}%
                    $\hss#1#2\hss$\crcr}}}\limits}
\def\brevefill#1#2{$\m@th\sbox\tw@{$#1($}%
  \hss\resizebox{#2}{\wd\tw@}{\rotatebox[origin=c]{90}{\upshape(}}\hss$}
\makeatletter

\title{Generative Modeling of Discrete Joint Distributions by \\ E-Geodesic Flow Matching on Assignment Manifolds}

\author[1]{Bastian~Boll}
\author[1]{Daniel~Gonzalez-Alvarado}
\author[1]{Christoph~Schnörr}
\affil[1]{%
    Institute for Mathematics\\
    Image and Pattern Analysis Group\\
    Heidelberg University, Germany
}
\begin{document}
\maketitle

% !TEX root =  ../AF_Generative-Discrete_uai2024.tex

\begin{abstract}
This paper introduces a novel generative model for discrete distributions based on continuous normalizing flows on the submanifold of factorizing discrete measures. Integration of the flow gradually assigns categories and avoids issues of discretizing the latent continuous model like rounding, sample truncation etc. General non-factorizing discrete distributions capable of representing complex statistical dependencies of structured discrete data, can be approximated by embedding the submanifold into a the meta-simplex of all joint discrete distributions and data-driven averaging. Efficient training of the generative model is demonstrated by matching the flow of geodesics of factorizing discrete distributions. Various experiments underline the approach's broad applicability.
\end{abstract}

%\noindent
%\textbf{ToDo:}
%\begin{itemize}
%\item Write about sampling (?) 
%\item Define $\exp, \log, \eins_{n}$
%\item In Section \ref{sec:likelihoods}: $\tilde v_t := {\log_{\eins_{\mc W}}}_{\#}%(\nu_t)$, s.t. $\tilde v_t$ has support in $T_0\mc W$, some notes in Appendix.
%\end{itemize}

\section{Introduction}\label{sec:introduction}
Generative models \cite{Kobyzev:2019aa,Papamakarios:2021vu} define an active area of research. They include, in particular, deep diffusion models \cite{Song:2021aa, Yang:2023aa}. The simple diffusion processes involved, however, lead to long training times and specialized algorithmic methods are required for efficient sampling.

As an alternative, the recent paper \cite{Lipman:2023} introduces the \textit{flow matching} approach to generative modeling which enables more stable and efficient training. This is achieved by taking as loss function the expected distance of a time-variant parametrized vector field to the vector field which generates the time-variant marginal distribution $p_{t},\,t\in[0,1]$, that connects a reference measure $p_{0}$ and the target distribution $p_{1}$. The key insight which makes the approach efficient is that the loss function gradients can be computed, by averaging over the sample set, the distance to the generating vector fields \textit{conditioned} on \textit{individual} data samples. The concrete form of the time-variant conditional measures generated in this way determine a whole class of flow-matching approaches and the form of the \textit{local} approximation of the data distribution. Even in the simplest case of basic Gaussian measures with time-variant parameters, the approach subsumes diffusion paths and hence provides an attractive alternative to current practice of diffusion-based approaches. More sophisticated paths by optimal Gaussian measure transport \cite{McCann:1997aa,Takatsu:2010aa} can be easily adopted and further improve the method.

Our approach is closely related to the recent extension of the flow matching approach to Riemannian manifolds \cite{Chen:2023}. It provides a generative model for \textit{discrete} or \textit{categorial} data, which defines another active subarea of research (cf.~Section \ref{sec:Related-Work}). This seamless combination of a Riemannian geometric structure for the respresentation and generation of discrete data, and the flow-matching approach for efficient and stable training, appears to be new in the literature. Our approach utilizes, in particular, the embedding approach of categorial distributions recently studied by  \cite{Boll:2023aa,Boll:2024aa}.

\begin{figure}[t]
    \centering
    \includegraphics[width=0.35\textwidth]{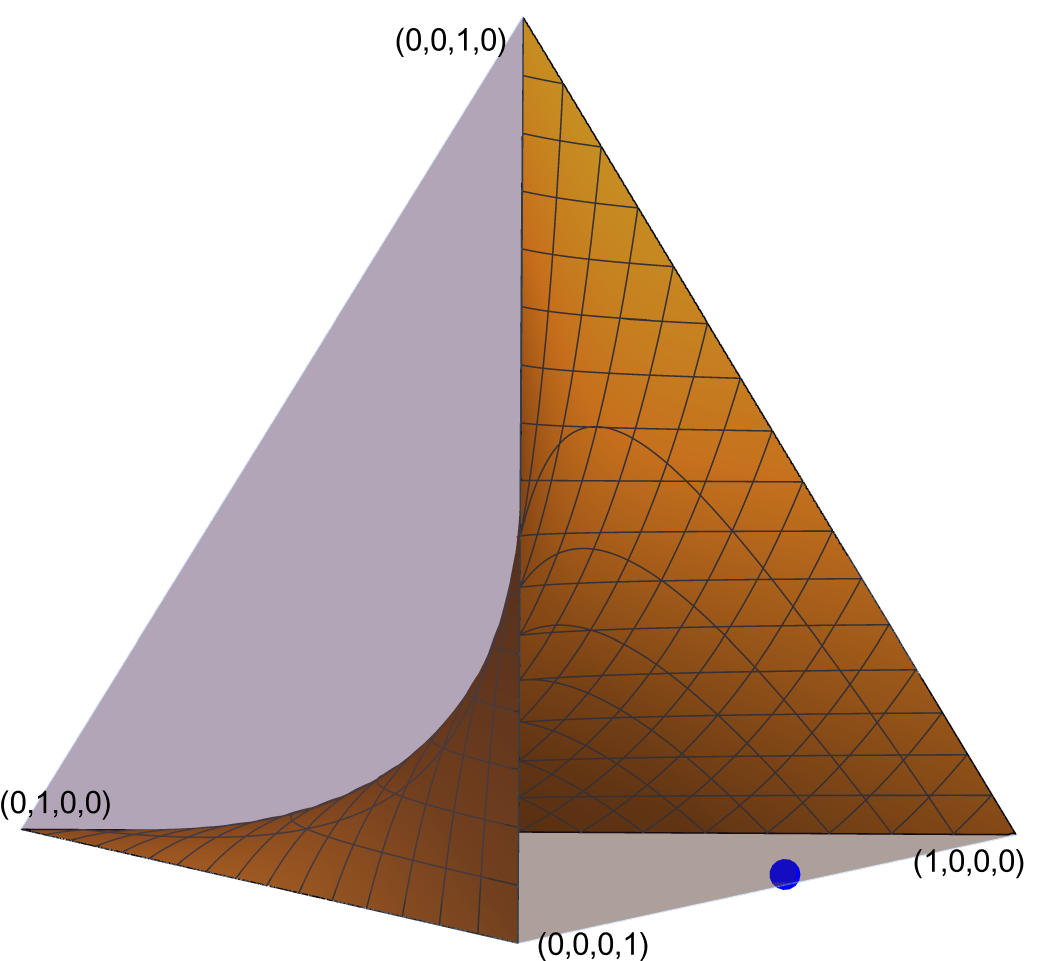}
    \caption{The tetrahedron represents in local coordinates all joint distributions $w\in\Delta^{4}$ (with $4=c^{n}$) of $n=2$ variables taking $c=2$ values. The embedded surface is the assignment manifold $\mc{W}$ of all factorizing distributions of 2 binary variables. The blue point represents a target joint distribution $p(y_{1},y_{2})=\frac{1}{100}(45,5,5,45)^{\T}$ with strong statistical dependency, i.e.~it is \textit{not} close to the \textit{any} factorizing distribution. This paper introduces a generative model for representing arbitrary discrete distributions as convex combination of hard category assignment distributions corresponding to the extreme points. Figure \ref{fig:Wright-AF} illustrates the representation of the target distribution (blue point).}
    \label{fig:Wright}
\end{figure}

\begin{figure}[t]
    \centering
    \includegraphics[width=0.35\textwidth]{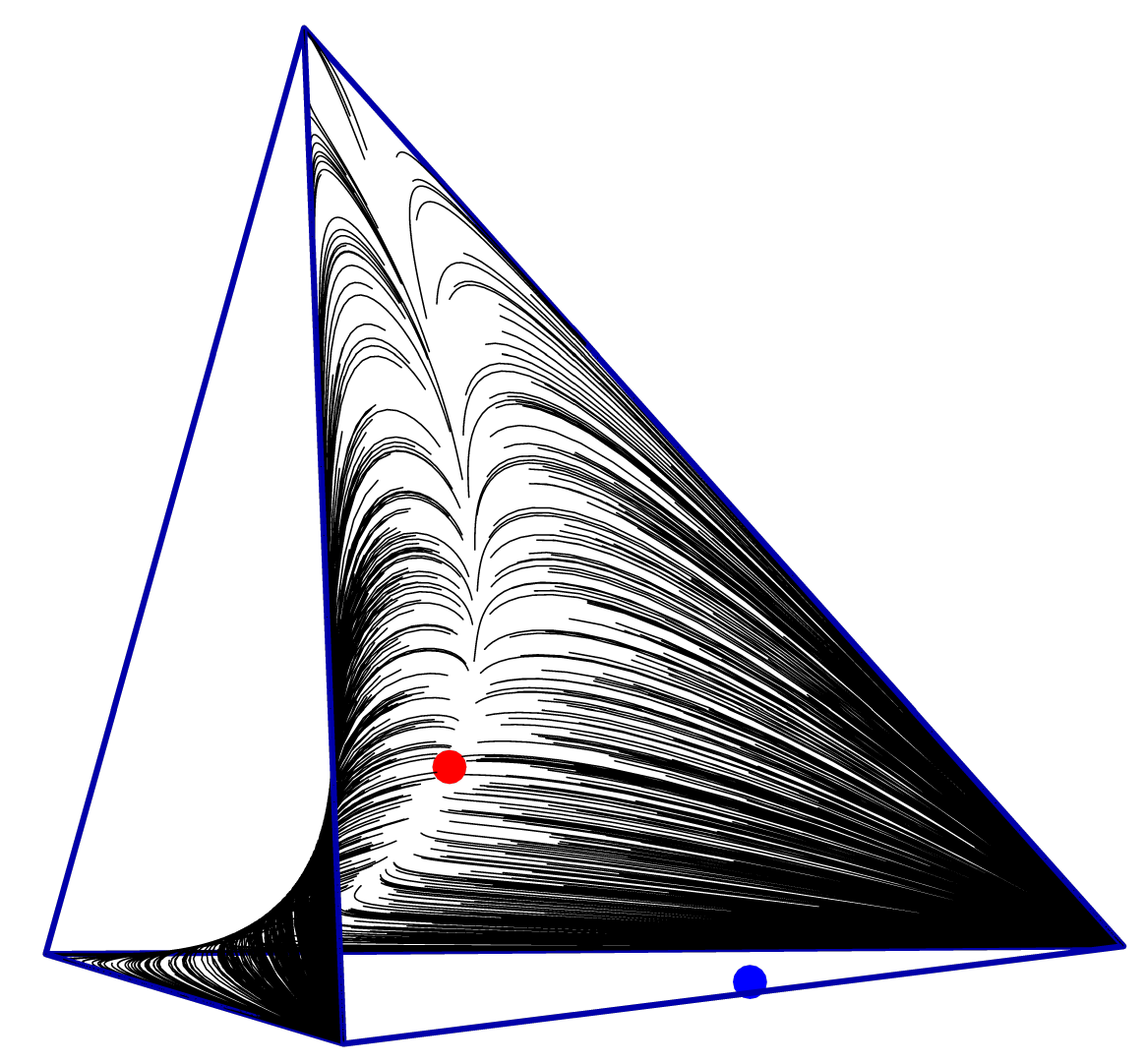}
    \caption{Visualization of 1000 samples from the target distribution (blue point; cf.~Figure \ref{fig:Wright}). Each sample corresponds to an integral curve $T(W(t))$ \eqref{eq:def-T-embedding} of the assignment flow ODE \eqref{eq:def-AF} on the embedded submanifold of factorizing distributions $\mc{W}\subseteq\mc{S}_{4}$, which can be computed efficiently by geometric integration. The entire assignment flow pushes forward a standard Gaussian reference distribution on the tangent space at the barycenter (red point), which is lifted to the submanifold and transported to the extreme points. The resulting `weights' represent the blue target distribution as convex combination. The parametrized vector field of the generative model is trained in a stable and efficient way by matching e-geodesic curves on the assignment manifold, which represent the training data and can be computed in closed form.
    }
    \label{fig:Wright-AF}
\end{figure}

Specifically, as illustrated for a toy problem by Figure \ref{fig:Wright}, we design and learn a dynamical system whose flow evolves on the submanifold of all \textit{factorizing} discrete distributions. This submanifold is embedded into the \textit{meta-simplex} of \textit{all} (possible) discrete distributions of $n$ discrete random variables, each of which takes $c$ values. Since the extreme points of the embedded submanifold and the ambient simplex coincide, pushing forward a basic reference measure -- as illustrated by Figure \ref{fig:Wright-AF} -- to (possibly a subset of) these extreme points, entails via the flow-matching training objective a convex combination of the extrem point measures, each of which represents a hard category assignment to each discrete random variable. As a result, we implicitly have a `universality property' in that (theoretically) \textit{any} discrete joint distribution can be represented and be sampled from using our approach.

\newpage
In summary, \textbf{this paper contributes}
\begin{itemize}
    \item a novel continuous normalizing flow model of discrete joint distributions; 
    \item a geometric Riemannian representation of time-variant push-forward measures which is efficient and stable learnable based on matching geodesic flows;
    \item a novel approach to data-driven approximations of general discrete distributions by submanifold embedding and averaging.
\end{itemize}

Section \ref{sec:Related-Work} reports related work on generative models on manifolds and for discrete distributions, respectively. Section \ref{sec:Background} specifies the dynamical systems tailored to discrete distributions and the embedding of the submanifold of factorizing distributions in the meta-simplex. Our approach is introduced in Section \ref{sec:Approach}. Experimental results are discussed in Section \ref{sec:Experiments}. We conclude in Section \ref{sec:Conclusion} and provide supplemental material in the appendix.

\section{Related Work}\label{sec:Related-Work}
We distinguish two areas of related research:
\begin{itemize}
    \item generative models on \textit{manifolds}, and
    \item generative models of \textit{discrete} distributions.
\end{itemize}

\subsection{Generative models on manifolds} Our work is inspired by and closely related to \cite{Chen:2023}. This work is based on the flow-matching approach to generative modeling introduced in the recent paper \cite{Lipman:2023}, which enables more stable and efficient training of continuous normalizing flows and hence provides an attractive alternative to maximum likelihood learning. The paper \cite{Chen:2023} extends the flow-matching approach to Riemannian manifolds and distinguishes two scenarios: \text{simple manifolds} on which geodesics can be computed in closed form, and general (non-simple) manifolds where approximate distances like the truncated diffusion distance are proposed instead.

Our approach adds the statistical (assignment) manifold $\mc{W}$ with the corresponding Fisher-Rao geometry to the list of simple manifolds in \cite{Chen:2023}. In addition, we propose a generative model for approximating joint distributions of \textit{discrete} random variables using the flow-matching approach, which essentially rests upon the embedding $T(\mc{W})\subset\mc{S}_{N}$ of the assignment manifold $\mc{W}$ in the meta-simplex $\mc{S}_{N}$ (Figures \ref{fig:Wright} and \ref{fig:approach}). To the best of our knowledge, our geometric approach to combining these three aspects appears to be novel in the literature.

Since our approach utilizes flow-matching of e-geodesics (i.e.~autoparallel curves with respect to the e-connection; cf.~\cite{Amari:2000aa}), which can be computed in closed form, it shares the advantages of generative models for distributions on other simple manifolds proposed so far, over related work using alternative methods \cite{Chen:2023}: significantly more stable and efficient training in comparison to using the maximum likelihood criterion (e.g., \cite{Lou:2020aa,Mathieu:2020aa}), better scalability to high dimensions than, e.g., \cite{Ben-Hamu:2022aa,Rozen:2021aa}, and no need for incorporating a costly simulation subroutine and approximation into the training procedure, as required in connection with extensions of diffusion-based generative models to manifolds (e.g., \cite{Huang:2022aa,De-Bortoli:2022aa}).

\subsection{Generative models of discrete distributions} \textit{Discrete distribution} means a probability distribution of a discrete random variable, i.e.~a random variable taking values in a finite set, or the joint distribution of several discrete random variables. Machine learning scenarios with discrete distributions have been studied from various angles in the literature.

The papers \cite{Maddison:2017aa,Jang:2017aa} study discrete random variables encoded by unit vectors, that are mollified via the softmax function, and sampling via the argmax operation perturbed by Gumbel-distributed noise (cf.~\cite{Hazan:2012aa}), both mainly motivated by applying the reparametrization trick and large-scale stochastic gradient via automatic differentiation, in connection with discrete random variables. A variant based on Gaussian distributions has been proposed by \cite{Potapczynski:2020aa}. This differs from our objective to represent and sample from complex discrete \textit{joint} distributions, which would be difficult to achieve using parametric densities on the probability simplex, like the Gumbel, Dirichlet and other distributions \cite{Aitchinson:1982vt}. 

Conversely, \cite{Tran:2019aa} apply a discrete change-of-variable formula and ensure invertibility of the proposed Discrete Flow architecture by modulo arithmetic of integers. This creates a highly non-smooth scenario which affects gradient-based training and apparently is not fully understood.

Accordingly, the authors of \cite{Lippe:2021aa} criticize the limited performance of discrete transformations \cite{Tran:2019aa,Hoogeboom:2019aa} regarding vocabulary size and gradient approximation. Likewise, dequantization techniques as proposed by \cite{Dinh:2017aa,Ho:2019aa}
 are limited regarding the representation of multidimensional statistical relations between categories. The normalizing flow approach proposed by \cite{Lippe:2021aa} uses a factorizing decoder, i.e.~the categorial variables are conditionally independent given the latent variables. Authors concede that this tends to limit model expressivity. 

Chen et al \cite{Chen:2022aa} propose a generative model for discrete distributions
$p(y)=p(y_{1},\dotsc,y_{n})$ of $n$ categorial variables, each taking $c_{i}$ values, based on continuous normalizing flows and quantisation of the generated probability distribution $p(x), x \in \R^{nD}$ (cf., e.g., \cite{Graf:2000aa},\cite{Gruber:2004aa}) whose geometry in terms of the Voronoi partition can be learned for each variable. An additional benefit of this approach is that for each cell a distribution $q(x|y_{i})$ is learned which enables dequantization $y\mapsto x$ after learning the discrete distribution $p(y)$. 
The dimension $D$ of the continuous latent domain ranges from $2 \dots 6$ in the paper and the best choice of $D$ and the numbers of Voronoi cells $N_{i},\; i\in [n]$, depending on the discrete distribution parameters $c_{i},\; i\in[n]$ and $n$, apparently is open. The ability to choose a small dimension $D$ independent of the number of categories, is an advantage. On the other hand, the representation of Voronoi cells by intersecting rays and linear inequality constraints is numerically subtle regarding automatic differentiation and training, and degenerate Voronoi cells may arise depending on the initial anchor points and the support of the underlying continuous distribution. Furthermore, maximum likelihood learning is employed which is less stable and efficient than the flow-matching approach \cite{Chen:2023aa} (cf.~Section \ref{sec:riemannian_flow_matching} below).

The paper \cite{Hoogeboom:2021aa} introduces Argmax Flows and Multinomial Diffusion which are closer in spirit to our approach in that learnable descision regions in a flat space are used for categorization. On the other hand, the ELBO variational bound \cite{Blei:2017aa} is applied together with Gaussian or Gumbel-thresholding for the argmax layer, which contrasts with our approach which uses a geodesic flow-matching objective function and learnable smooth regions on the tangent bundle defining category assignment.

Finally, we point out the very recent paper \cite{Chen:2023aa} which uses a real-valued representation of bit-encoded discrete or categorial data, in connection with diffusion-based generative models. Impressive empirical results are reported. We refer again, however, to the superiority of the flow-matching approach relative to diffusion models \cite{Lipman:2023},\cite{Chen:2023}.

In view of the two paragraphs above, we point out that the generative model introduced in this paper seamlessly combines a flow-matching approach with a Riemannian geometric structure tailored to represent discrete distributions and to generate discrete and categorial data.

\section{Background}\label{sec:Background}
\subsection{Assignment Flows}\label{sec:background-AF}
Denote by $\mc{S}_{c}:=\mathring{\Delta}_{c} = \{p\in\mathbb{R}^{c}\colon p_{j}>0,\,\langle\mathbb{1}_{c},p\rangle=1,\,\forall j\in[c]\}$ the relative interior of the probability simplex $\Delta_{c}$, i.e.~the set of discrete probability distributions of categorial random variables $y$ which take values in the set $[c]=\{1,2,\dotsc,c\},\,c\in\mathbb{N}$. $\mc{S}_{c}$ equipped with the Fisher-Rao metric $g_{w}(u,v)=\la u,\Diag(w)^{-1} v\ra,\, u,v\in T_{0}$, is a Riemannian manifold, with the trivial tangent bundle $\mc{S}_{c}\times T_{0}$ and tangent space $T_{0}=\{v\in\R^{c}\colon \la\eins_{c},v\ra=0\}$.

The \textit{assignment manifold} $\mc{W}=\mc{S}_{c}\times\dotsb\times\mc{S}_{c}\subset\R_{\geq}^{n\times c}$ ($n$ factors) is a corresponding product manifold with trivial tangent bundle $\mc{W}\times\mc{T}_{0}$ and tangent space $\mc{T}_{0}=T_{0}\times\dotsb\times T_{0}$, equipped with the product Fisher-Rao metric $g_{W}(U,V)=\sum_{i\in[n]}g_{W_{i}}(U_{i},V_{i}),\,W\in\mc{W},\, U,V\in\mc{T}_{0}$. Each point $W = (W_{1},\dotsc,W_{n})^{\T}\in\mc{W}$ represents a \textit{factorizing} joint distribution of $n$ discrete random variables $y_{1},\dotsc,y_{n}$, each taking values in $[c]$.

\textit{Assignment flows} denote a class of dynamical systems of the form
\begin{equation}\label{eq:def-AF}
    \dot W(t) = R_{W(t)}\big[F_{\theta}\big(W(t)\big)\big],\; W(0)=W_{0}\in\mc{W}
\end{equation}
where the generating vector field on the right-hand side is given by a parametrized function $F_{\theta}\colon\mc{W}\to\R^{n\times c}$ with parameters $\theta$, and $R_{W}$ is a $\mc{T}_{0}$-valued linear map
\begin{subequations}\label{eq:def-R}
    \begin{align}
        \big(& R_{W}[F_{\theta}(W)]\big)_{i}
        = R_{W_{i}}F_{\theta; i}(W),\quad i\in[n]
        \\
        &= \Diag(W_{i}) F_{\theta; i}(W)
        - \big\la W_{i}, F_{\theta; i}(W)\big\ra,
    \end{align}
\end{subequations}
which represents the inverse metric tensor in the coordinates of the ambient Euclidean space $\R^{n\times c}$. Given the initial point $W_{0}$ in general position, which encodes structured data in concrete applications, $\lim_{t\to\infty}W(t) = (e_{j_{1}},\dotsc,e_{j_{n}})^{\T}$ converges to an extreme point of the closure of $\ol{\mc{W}}$, which corresponds to an hard label assignment for each discrete variable $y_{i} = j_{i}\in[c],\, i\in[n]$ in terms of the canonical unit vectors $e_{j_{i}}\in\R^{c}$. Assignment flows have been introduced in \cite{Astrom:2017ac} with basic properties (well-posedness, convergence) established in \cite{Zern:2020aa} and a wide range of efficient geometric integration schemes for computing $W(t)$ \cite{Zeilmann:2020aa}. 

The exponential map with respect to the e-connection reads
\begin{equation}\label{eq:def-Exp}
    \Exp_{p}(v) = \frac{p\cdot e^{\frac{v}{p}}}{\la p, e^{\frac{p}{v}}\ra},\qquad p\in\mc{S}_{c},\; v\in T_{0},
\end{equation}
where multiplication $\cdot$ and the exponential function apply componentwise. We set
\begin{subequations}\label{eq:def-exp}
\begin{align}
    \exp_{p}&\colon T_{0}\to\mc{S}_{c},\qquad p\in\mc{S}_{c},
    \\
    \exp_{p} &:= \Exp_{p}\circ R_{p}.
\end{align}
\end{subequations}
Both mappings \eqref{eq:def-Exp} and \eqref{eq:def-exp} extend factor-wise to the product space $\mc{T}_{0}=T_{0}\mc{W}$ analogous to \eqref{eq:def-R}. 
For more details on information geometry, we refer to \cite{Amari:2000aa}.
%%%
\subsection{Meta-Simplex Flow Embedding}\label{subsec:embedding}
The assignment manifold $\mc{W}$ only represents \textit{factorizing} distributions which forms a very specific subset of \textit{all} joint distributions of $n$ discrete random variables $y_{1},\dotsc,y_{n}$, each taking $[c]$ values. Indeed, a \textit{general} distribution is specified by the combinatorially large number of $N := c^{n}$
values $p(y_{1},\dotsc,y_{n})$ of the joint probability distribution, whereas a point $W\in\mc{W}$ on the assignment manifold merely has $n\cdot c$ coordinates. In order to approximate general discrete distributions which typically are supposed to represent complex statistical dependencies of discrete structured output, we introduce the corresponding \textit{meta-simplex} of \textit{all} discrete distributions
\begin{equation}\label{eq:def-SN}
\mc{S}_{N} = \big\{p\in\R_{>}^{N}\colon\la\eins_{N},p\ra=1,\;\forall j\in[N]\big\}.
\end{equation}
For example, the joint distribution of two binary variables ($n=c=2$) is a point $p(y_{1},y_{2})\in\mc{S}_{4}$. By contrast, if $p(y_{1},y_{2})=W\in\mc{W}\subset\mc{S}_{4}$ is a distribution on the assignment manifold, then $W = (\bsm w_{1} \\ 1-w_{1} \esm, \bsm w_{2} \\ 1-w_{2} \esm)^{\T}$. The corresponding two-dimensional submanifold $\mc{W}$ embedded in $\mc{S}_{4}$ is depicted by Figure \ref{fig:Wright}. From the viewpoint of mathematics, such embedded sets are known as \textit{Segre varieties} at the intersection of algebraic geometry and statistics \cite{Lin:2009aa, Drton:2009aa}. The corresponding embedding map reads
\begin{equation}\label{eq:def-T-embedding}
T\colon\mc{W}\to\mc{T}\subset\mc{S}_{N},\quad
T(W)_{\alpha} := \prod_{i\in[n]} W_{i,\alpha_{i}},
\end{equation}
with the multi-index notation $\alpha=(\alpha_{1},\dotsc,\alpha_{n})\in [c]^{n}$. For the above example with $n=c=2$, one has $T(W)=(w_{1} w_{2}, w_{1}(1-w_{2}), (1-w_{1}) w_{2}, (1-w_{1})(1-w_{2}))^{\T}$. The mapping \eqref{eq:def-T-embedding} has been introduced and studied recently in \cite{Boll:2023aa,Boll:2024aa}. Each multi-index $\alpha$ indexes exactly one extreme point $e_{\alpha}\in\{0,1\}^{N}$ (unit vector, discrete Dirac measure) of the meta-simplex $S_{N}$, and we identify accordingly $\alpha$ with this discrete extreme measure, which represents a hard category or \textit{label assignment} to each discrete random variable $y_{1},\dotsc,y_{n}$. We therefore call $\alpha$ also \textit{label configuration}.

The components of a point $p\in\mc{S}_{N}$ in \eqref{eq:def-SN} are indexed by $\alpha$ as well, analogous to the embedded vectors \eqref{eq:def-T-embedding}. We use the short-hand
\begin{equation}
    p(\alpha) := p(y_{\alpha_{1}},\dotsc,y_{\alpha_{n}}) = p_{\alpha},\quad\alpha\in[c]^{n}.
\end{equation}

The following proposition highlights the specific role of the embedded assignment manifold $T(\mc{W})\subset\mc{S}_{N}$.
\\[0.2cm]
\textbf{Proposition 1}(\cite[Prop.~3.2]{Boll:2024aa})
For every $W\in\mc{W}$, the distribution $T(W)\in\mc{S}_{N}$ has maximum entropy 
\begin{equation}
H\big(T(W)\big) = -\sum_{\alpha\in[c]^{n}} T(W)_{\alpha}\log T(W)_{\alpha}
\end{equation}
among all $p\in\mc{S}_{N}$ subject to the marginal constraint 
$M p = W$, 
where the marginalization map $M\colon\R^{N}\to\R^{n\times c}$ is given by
\begin{equation}\label{eq:def-M}
(M p)_{i,j} := \sum_{\alpha\in[c]^{n}\colon \alpha_{i}=j} p_{\alpha},\;
\forall (i,j)\in [n]\times [c].
\end{equation}
Any general distribution $p\in\mc{S}_{N}\setminus\mc{W}$ which is \textit{not} in $T(\mc{W})$ has \textit{non}-maximal entropy and hence is \textit{more} informative by encoding additional statistical dependencies \cite{Cover:2006aa}. Our approach for generating general distributions $p\in\mc{S}_{N}$, by combining simple distributions $W\in\mc{W}$ via the embedding \eqref{eq:def-T-embedding} and assignment flows \eqref{eq:def-AF}, is introduced next.

\section{Approach}\label{sec:Approach}
\begin{figure}
    \centering
    \includegraphics[width=0.45\textwidth]{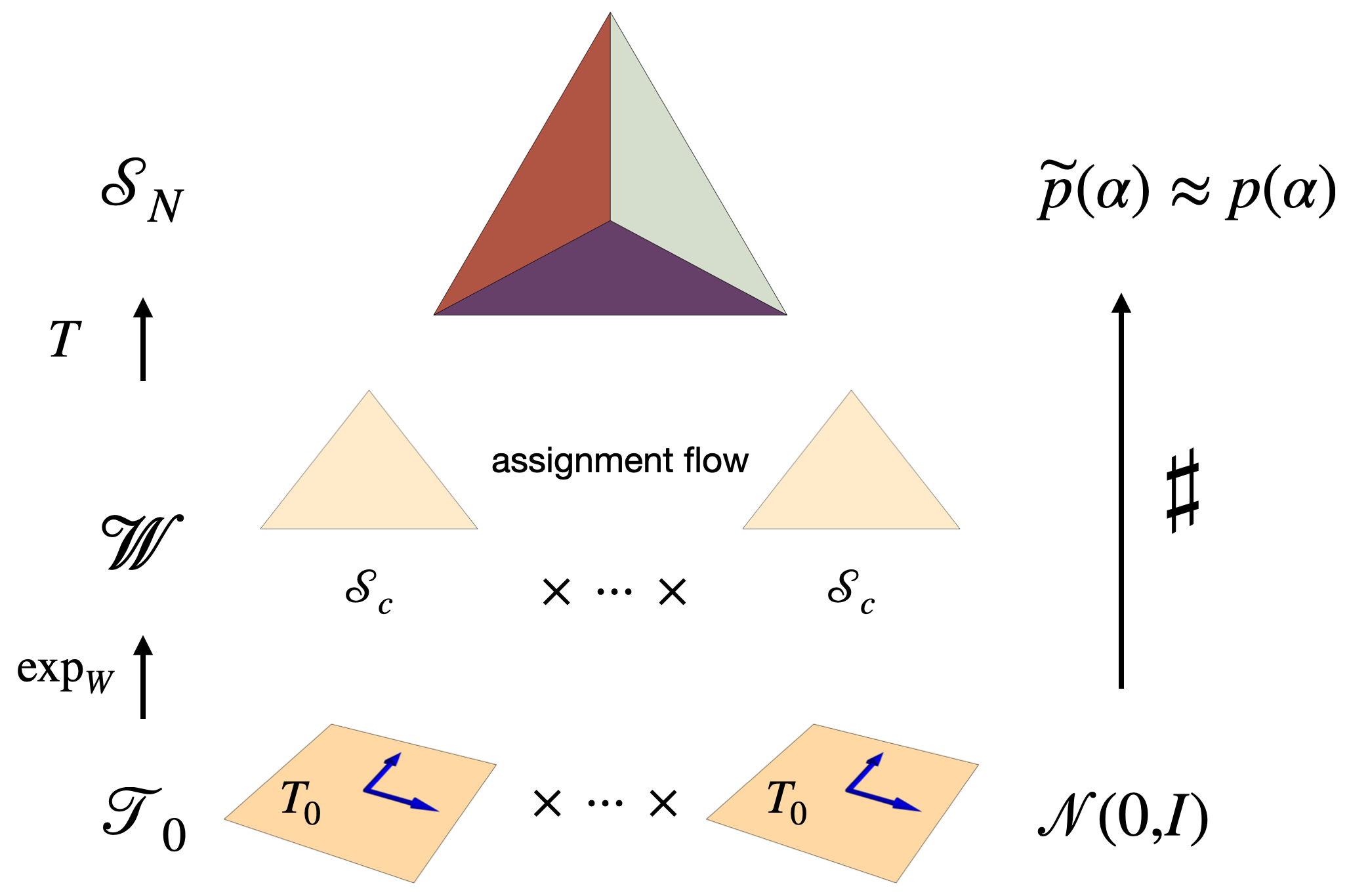}
    \caption{\textbf{Overview of the approach:} The standard Gaussian reference measure $\mc{N}(0,I)$ is pushed forward by the exponential map $\exp_{W}$ from the flat tangent product space $\mc{T}_{0}$ to the assignment manifold $\mc{W}$, and further to the meta-simplex $\mc{S}_{N}$ \eqref{eq:def-SN} by geometrically integrating the assignment flow \eqref{eq:def-AF}. Since the assignment flow converges to the extreme points of $\ol{\mc{W}}$ which agree with the extreme points of $\mc{S}_{N}$, an approximation $\wt{p}(\alpha)$ of a general \textit{discrete} target measure $p(\alpha)$ underlying given data can be approximated by matching the flow of e-geodesics (corresponding to data samples) and convex combination in terms of embedded factorized distributions $T(W),\,W\in\mc{W}$ and empirical expectation.}
    \label{fig:approach}
\end{figure}

Every joint distribution $p\in \mc{S}_N$ can be written as convex combination of extreme point measures
\begin{equation}\label{eq:data_distr_convex}
	p(\alpha) = \big(\sum_{\beta\in [c]^n} p_\beta e_\beta\big)_\alpha = \EE_{\beta\sim p}[T(M e_\beta)]_\alpha,
\end{equation}
where the right-hand side denotes the expectation of the argument with respect to $p$. 
We will employ flow-matching, which approximates the data distribution by a mixture of simple conditional distributions $q(\alpha|\beta)$
\begin{equation}\label{eq:data_cond_mixture}
	p(\alpha) \approx \int q(\alpha|\beta)p(\beta)d\beta = \EE_{\beta\sim p}[q(\alpha|\beta)] =: \wt p(\alpha).
\end{equation}
Comparing \eqref{eq:data_cond_mixture} and \eqref{eq:data_distr_convex} motivates the Ansatz
\begin{equation}\label{eq:discrete_flow_matching_ansatz}
	q(\alpha|\beta) := \EE_{W\sim \nu_\beta}[T(W)_\alpha] \approx T(M e_\beta)_\alpha,
\end{equation}
where the latter approximation holds 
for a simple distribution $\nu_\beta$ on $\mc{W}$ concentrated close to the extreme point $M e_\beta$ on $\mc{W}$, which corresponds (cf.~Eq.~\eqref{eq:def-M}) to the extreme point $e_{\beta}\in\mc{S}_{n}$. The distribution $q(\alpha|\beta)$ defined by \eqref{eq:discrete_flow_matching_ansatz} is \textit{simple}, because the right-hand side merely involves a factorizing distribution $W\in\mc{W}$ on the embedded assignment manifold, the embedding map $T$ \eqref{eq:def-T-embedding} and averaging.

Figure \ref{fig:approach}, center and top row, illustrates this part of the approach.

\subsection{Riemannian Flow Matching}\label{sec:riemannian_flow_matching}
A distribution $\nu$ on $\mc{W}$ can approximately represent unknown discrete data distributions $p\in \mc{S}_N$ through
\begin{equation}\label{eq:joint_representation}
	p_\alpha \approx \EE_{W\sim \nu}[T(W)_\alpha].
\end{equation}
The underlying idea is that every distribution $p\in \mc{S}_N$ can be represented as a mixture of Dirac distributions on extreme points \eqref{eq:data_distr_convex}. 
\cite{Chen:2023} describe a method for learning continuous normalizing flows on Riemannian manifolds via flow matching. In our case, the manifold in question is the assignment manifold $\mc{W}$ and we are learning a measure $\nu$ on $\mc{W}$ to represent a discrete data distribution $p$ via \eqref{eq:joint_representation}.
A core contribution of \cite{Chen:2023} was to show how flow matching to learn $\nu$ can be achieved by matching \textit{conditional} probability distributions $\nu_\beta:=\nu(W|\beta)$ for data samples $\beta\sim p$.
This corresponds to the distribution $\nu_\beta = \delta_{\epsilon\eins_\mc{W} + (1-\epsilon)Me_\beta}\in\mc{W}$ in the ansatz \eqref{eq:discrete_flow_matching_ansatz}, with $0<\epsilon\ll 1$ and the barycenter $\eins_{\mc{W}}$ of $\mc{W}$. $\nu_{\beta}$ is closely concentrated around the extreme point of $\ol{\mc{W}}$ corresponding to the extreme point $e_{\beta}\in\mc{S}_{N}$.

We now describe how $\nu$ is learned via a normalizing flow and geodesic flow matching, as illustrated by the bottom part of Figure \ref{fig:approach}. 
Fix $\epsilon$ and define the representation 
\begin{equation}\label{eq:smoothed_corner}
	q_\beta = \epsilon\eins_\mc{W} + (1-\epsilon)Me_\beta \in\mc{W}
\end{equation}
of sample data $\beta\sim p$ on $\mc{W}$. 
Define the reference distribution 
\begin{equation}\label{eq:reference_distr}
	\nu_0 = (\exp_{\eins_\mc{W}})_\sharp\mc{N}_0
\end{equation}
on $\mc{W}$ as push-foward with respect to the exponential map of the standard Gaussian $\mc{N}_{0}=\mc{N}(0,I)$ centered in the tangent space at $0\in\mc{T}_{0}$, and the define conditional distributions 
\begin{equation}\label{eq:rfm_endpoints}
	\nu_0(W|q_\beta) = \nu_0(W),\qquad \nu_1(W|q_\beta) = \delta_{q_\beta}
\end{equation}
as endpoints of a probability path defined below, 
for each configuration $\beta\in [c]^n$. Clearly, one has 
\begin{subequations}\label{eq:rfm_endpoint_validity}
\begin{align}
    \EE_{q_\beta\sim \nu}[\nu_0(W|q_\beta)] &= \nu_0(W),\\
 	\EE_{q_\beta\sim \nu}[\nu_1(W|q_\beta)] &= \EE_{q_\beta\sim \nu} [\delta_{q_\beta}] = \nu,
\end{align}
\end{subequations} 
due to the unconditional reference measure in the former case and the Dirac measure in the latter. 
As a consequence, we can flow-match the \textit{probability path}
\begin{equation}\label{eq:rfm_prob_path}
	\nu_t(W) = \EE_{q_\beta\sim \nu}[\nu_t(W|q_\beta)]
\end{equation}
by merely flow-matching \textit{conditional} probability paths $\nu_t(W|q_\beta)$ with endpoints \eqref{eq:rfm_endpoints} for \textit{individual} data samples $\beta\sim p$. A simple choice is the probability path generated by pushing $\nu_0$ along $e$-geodesics connecting each initial $W_{0}\in \mc{W}$ governed by $\nu_{0}$  with $q_\beta$. 

Denote a latent tangent vector on $\mc{T}_{0}$ by $u_0\sim \mc{N}_0$ and the corresponding point on $\mc{W}$ by $W_0 = \exp_{\eins_\mc{W}}(u_0)$. Further, denote by $u_\beta = \exp_{\eins_\mc{W}}^{-1}(q_\beta)\in \mc{T}_{0}$ the tangent vector corresponding to $q_\beta$. Then the path of an $e$-geodesic connecting $W_0$ with $q_\beta$ reads
\begin{subequations}\label{eq:cond_prob_path_exp}
\begin{align}
	W_t^\beta &= \exp_{W_0}(t\exp_{W_0}^{-1}(q_\beta))\\
		%&= \exp_{\eins_\mc{W}}(v_0 + t\Pi_0\log \frac{q_\beta}{W_0})\\
		&= \exp_{\eins_\mc{W}}(u_0 + t(u_\beta - u_0))
\end{align}
\end{subequations}
and differentiation gives
\begin{equation}\label{eq:cond_prob_diff_exp} 
	\dot W_t^\beta = R_{W_t^\beta}[u_\beta - u_0]
\end{equation}
with the linear map $R_{W_t^\beta}$ given by \eqref{eq:def-R} and factor-wise extension.  
The \textit{Riemannian conditional flow matching objective}, in the present case with the norm induced by the Fisher-Rao metric, reads
\begin{equation}\label{eq:rcfm_obj}
	\mc{L}_\text{RCFM}(\theta) = \EE_{t, q_\beta, W_0}\big[\|R_{W_t^\beta}[F_\theta(W_t^\beta)] - \dot W_t^\beta\|_{W_t^\beta}^2\big],
\end{equation}
It was shown in \cite{Lipman:2023, Chen:2023}, that \eqref{eq:rcfm_obj} has the same gradient with respect to parameters $\theta$ as the \emph{unconditional} Riemannian flow matching objective
\begin{equation}\label{eq:rfm_obj}
	\mc{L}_\text{RFM}(\theta) = \EE_{t, W_0}\big[\|R_{\psi_t}[F_\theta(\psi_t)] - U(\psi_t))\|_{\psi_t}^2\big],
\end{equation}
for a vector field $U$ which transports $\nu_0$ to $\nu$ and generates the assignment flow $\psi_t$ by integrating \eqref{eq:def-AF}.
Here, the expectation is taken with respect to $t\sim \mc{U}[0,1], q_\beta \sim \nu, W_0\sim \nu_0$ respectively and $\psi_t = \psi_t(W_0)$.

As \eqref{eq:cond_prob_diff_exp} suggests in view of \eqref{eq:def-AF}, we have written the parameterized vector field in \eqref{eq:rcfm_obj} as an assignment flow vector field with parameterized fitness function $F_\theta$.
Substituting \eqref{eq:cond_prob_diff_exp} into \eqref{eq:rcfm_obj} finally gives
\begin{equation}\label{eq:expandied_cond_obj}
%\begin{subequations}\label{eq:expandied_cond_obj}
%\begin{align}
	\mc{L}_\text{RCFM}(\theta) = \EE_{t, q_\beta, W_0}\big[\|R_{W_t^\beta}[F_\theta(W_t^\beta) -(v_\beta - v_0)]\|_{W_t^\beta}^2\big] %\\
%	&= \EE_{t, q_\beta, W_0}\la F_\theta(W_t^\beta) - (v_\beta - v_0), R_{W_t^\beta}[F_\theta(W_t^\beta) - (v_\beta - v_0)]\ra
%\end{align}
%\end{subequations}
\end{equation}
which we can directly use as a training objective to learn parameters $\theta$. Learning $\theta$ amounts to learning a flow \eqref{eq:def-AF} which generates $\nu$ through pushfoward of $\nu_0$ and approximates the data distribution by \eqref{eq:joint_representation}.

\subsection{Likelihood Computation}\label{sec:likelihoods}
For any configuration $\alpha\in [c]^n$, let $r_\alpha\subseteq \mc{W}$ denote the set of points on the assignment manifold which have $M e_{\alpha}$ as closest extremal point of $\ol{\mc{W}}$. 
%Sampling from 
We aim to bound the probability of $r_\alpha$ under the pushforward distribution $\nu_t = (\psi_t)_\sharp\nu_0$ from below. To this end, we construct a subset $\wt r_\alpha\subseteq \mc{T}_{0}$ such that 
\begin{equation}\label{eq:wt_r_constraint}
    \exp_{\eins_\mc{W}}(\wt r_\alpha)\subseteq r_\alpha
\end{equation}
Let $\wt \nu_t = (\exp_{\eins_\mc{W}}^{-1})_\sharp\nu_t$. Then
\begin{equation}\label{eq:likelihood_bound}
\log p_\alpha = \log \PP_{W\sim \nu_t}(r_\alpha) \geq \log \PP_{v\sim \wt \nu_t}(\wt r_\alpha).
\end{equation}
Because we learn the flow $\psi_t$ such that it concentrates probability mass close to points $q_\alpha$ (see \eqref{eq:smoothed_corner}), we can strategically choose $\wt r_\alpha$ to make importance sampling for this integral cheap to compute. Let $\rho$ be a proposal distribution with support $\wt r_\alpha$. Then
\begin{subequations}
\begin{align}
    \log\, \PP_{v\sim \wt \nu_t}(\wt r_\alpha) 
    &= \log \int_{\wt r_\alpha} \wt \nu_t(v)dv\label{eq:likelihood_exact_int}\\
        &= \log \int_{\wt r_\alpha} \frac{\wt \nu_t(v)}{\rho(v)}\rho(v) dv \\
        &= \log \EE_{v\sim \rho} \Big[\frac{\wt \nu_t(v)}{\rho(v)}\Big] \\
        &\geq \EE_{v\sim \rho} [\log \wt \nu_t(v) - \log \rho(v)]\label{eq:likelihood_int}
\end{align}
\end{subequations}
by Jensen's inequality and we can thus compute a lower bound on the log-likelihood of $\alpha$ under our model by approximating the integral in \eqref{eq:likelihood_int}.
Evaluating the integrand amounts to computing log-likelihood under a continuous normalizing flow. This can be done through a instantaneous change of variables \cite{Grathwohl:2018} and evaluating the log-likelihood under the proposal distribution, which we are able to do in closed form.
Since $\nu_t$ is trained to transport the reference distribution to one which is concentrated on points $q_\alpha$, we expect the integrand in \eqref{eq:likelihood_exact_int} to have most of its mass close to $\wt q_\alpha = \exp_{\eins_\mc{W}}^{-1}(q_\alpha)$, which motivates the following choice of $\wt r_\alpha$ and $\rho$.
Let $\wt r_\alpha$ be a sphere with diameter $d > 0$ centered at $\wt q_\alpha$ and choose $d$ small enough to satisfy \eqref{eq:wt_r_constraint}.
Let $\rho$ be an isotropic normal distribution with variance $\sigma^2$ and mean $\wt q_\alpha$ supported only on the sphere $\wt r_\alpha$.
The tail probability of this distribution required to normalize it to probability mass $1$ is analytically available because $\rho$ has rotational symmetry (see Appendix~\ref{sec:importance_sampling_computations}).

\section{Experiments}\label{sec:Experiments}

\begin{figure*}[ht]
	\centering
	\begin{subfigure}{\textwidth}
		\includegraphics[width=\textwidth]{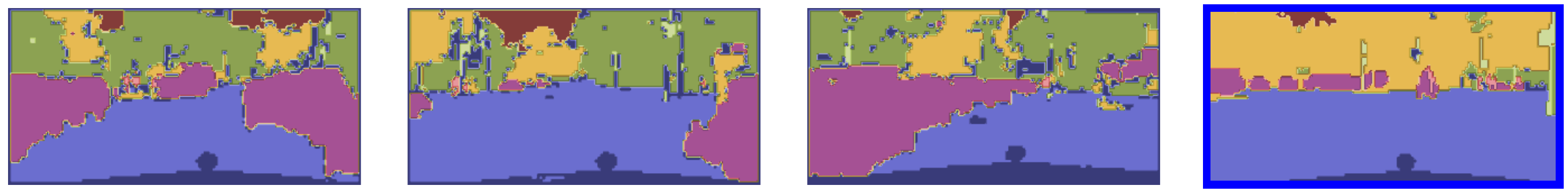}
		%\caption{SubFig 1}
	\end{subfigure}
%%%%%%%%%%%%%%
	\begin{subfigure}{\textwidth}
		\includegraphics[width=\textwidth]{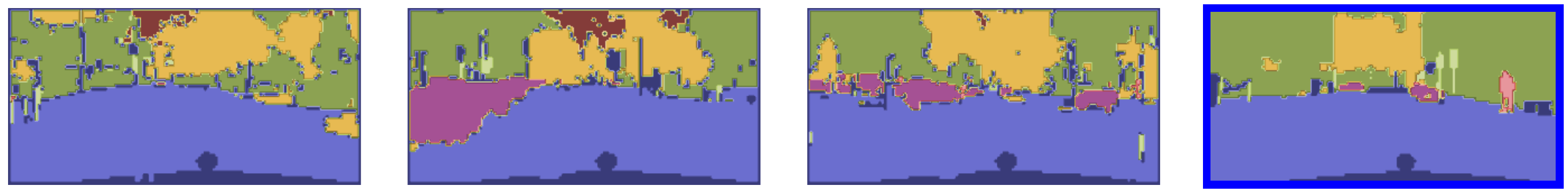}
	\end{subfigure}
%%%%%%%%%%%%%%
	\caption{Left: Random samples drawn from our model trained on discrete Cityscapes segmentation data ($c=8$ classes) at resolution $128\times 256$. Right with blue border: Randomly drawn training data.}
     \label{fig:sampling_cityscapes_reference}
\end{figure*}

\begin{figure*}
    \centering
    \includegraphics[width=\linewidth]{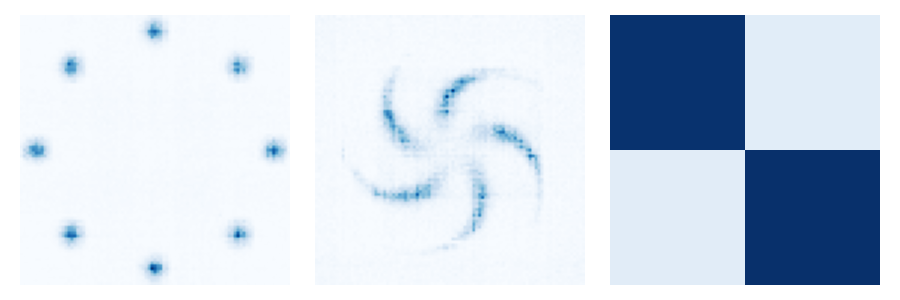}
    \caption{Histogram of samples from our model fitting the joint distribution of $n=2$ discrete random variables. Left and middle: $c=91$ classes per variable. Right: $c=2$ classes per variable. All three plots show values of the \emph{joint distribution}. Clearly, the model is able to fit multi-modal joint distributions which do not factorize into independent marginals. The plot on the right is the joint distribution shown as blue dot in Figures~\ref{fig:Wright} and~\ref{fig:Wright-AF}.}
    \label{fig:simple_distr}
\end{figure*}

\begin{figure}
    \centering
    \includegraphics[width=\linewidth]{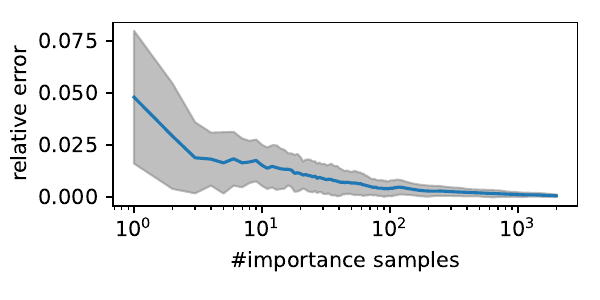}
    \caption{Convergence of importance sampling the integral \eqref{eq:likelihood_int} for our generative model of MNIST data. Already for few samples, relative error is within few percentage points. Mean and standard deviation are evaluated over $30$ data drawn at random from the MNIST test set.}
    \label{fig:sampling_convergence}
\end{figure}

\begin{figure}
    \centering
    \includegraphics[width=\linewidth]{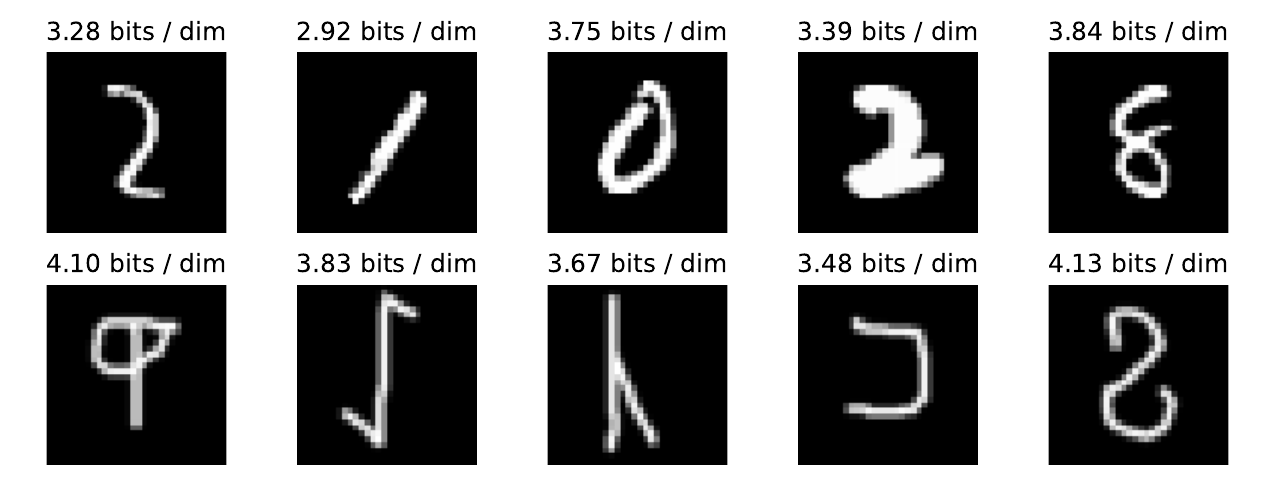}
    \caption{Upper-bound on the likelihood of samples from MNIST (in distribution, first row) and Omniglot (out of distribution, second row) under our model. Data with low likelihood (high bits/dim) can be detected as outliers. Likelihood was bounded by approximating \eqref{eq:likelihood_int} through $200$ importance samples.}
    \label{fig:outliers}
\end{figure}

\subsection{Generating Image Segmentations}\label{sec:experiments_img}
In image segmentation, a joint assignment of classes to pixels is usually sought conditioned on the pixel values themselves. Here, we instead focus on the \emph{unconditional} discrete distribution of segmentations, without regard to the original pixel data. These discrete distributions are very high-dimensional in general, with $N = c^n$ increasing exponentially in the number of pixels.
We perform Riemannian flow-matching \eqref{eq:rfm_obj} via the conditional objective \eqref{eq:rcfm_obj} to lean assignment flows \eqref{eq:def-AF} which approximate this discrete distribution via \eqref{eq:discrete_flow_matching_ansatz}. To this end, we parametrize $F_\theta$ by the UNet architecture of \cite{Dhariwal2021diffusion} (details in Appendix~\ref{app:experiment_details}) and train on the segmentations of Cityscapes \cite{Cordts2016Cityscapes}, downsampled to $c=8$ classes and resolution $128\times 256$, as well as MNIST \cite{lecun2010mnist}, regarded as binary $c=2$ segmentations of $28\times 28$ pixel images.

Figure~\ref{fig:sampling_cityscapes_reference} shows samples from the learned distribution of Cityscapes segmentations randomly drawn from our model.

\subsection{Numerical Likelihood Bounds}\label{sec:experiments_loglh}
In general, sampling integrals over high-dimensional domains is a difficult task. We evaluate the effectiveness of our importance sampling approach of Section~\ref{sec:likelihoods} empirically by plotting the relative error of likelihood bounds computed from varying number of importance samples. The reference value for the integral is computed from $3557$ samples. Figure~\ref{fig:sampling_convergence} shows that even a small number of samples suffices to compute a reasonable approximation to the integral. This is evaluated on MNIST (padded to size $32\times 32$) i.e. the domain of interation is $(c-1)n = 1024$ dimensional.

\subsection{Detecting out-of-distribution data}\label{sec:experiments_ood}
We test if the bound on log-likelihood \eqref{eq:likelihood_int} under our generative model allows to detect out-of-distribution data.
To this end, we train a model \eqref{eq:def-AF} to approximate the MNIST data distribution and compute the bound \eqref{eq:likelihood_int} by importance sampling.
The Omniglot dataset \cite{Lake2015} contains visually similar images to MNIST, but displays a wide variety of different symbols. Figure~\ref{fig:outliers} shows random samples of MNIST and Omniglot as well as the bound on log-likelihood of each sample under our learned MNIST distribution. 
Samples from this model can be found in Appendix~\ref{app:experiment_details}.

\section{Discussion}\label{sec:Discussion}

Riemannian flow matching on the assignment manifold is a remarkably stable and efficient approach to learning high-dimensional discrete distributions. 
Because training does not require simulation of flow trajectories, it requires few function evaluations compared to established likelihood-based normalizing flow training. 

To the best of our knowledge, our work is the first to demonstrate flow-matching for discrete data.
The Cityscapes experiment illustrates the resulting gain in efficiency. While the related work \cite{Hoogeboom:2021aa} studies the same discrete labeling dataset, they subsample to low resolution $32\times 64$ in order to save on computation.
In contrast, Figure~\ref{fig:sampling_cityscapes_reference} shows convincing samples at resolution $128\times 256$ from our model trained within 3.5 hours on a single desktop GPU.

\textbf{Current limitations}. Even though the importance sampling approach proposed in Section~\ref{sec:likelihoods} is empirically very effective, it still requires more computational effort to evaluate compared to typical normalizing flow models. This is because the integrand needs to be sampled by performing instantaneous change of variables and integrating the flow backward in time multiple times to get a precise and reliable estimate (Fig~\ref{fig:sampling_convergence}). Beyond computational considerations, there is also a Jensen gap in the estimate \eqref{eq:likelihood_int} which can not be reduced by drawing more samples.
%For the low-dimensional example 
There is some potential to make \eqref{eq:likelihood_bound} an equality in the future by choosing $\wt r_\alpha$ accordingly.
In principle, one can also consider direct simulation of \eqref{eq:joint_representation} as an alternative which estimates exact likelihood without the Jensen gap. A numerically stable procedure to perform this sampling was suggested in \cite{Boll:2023aa}, but the approach still only works for low dimensions in practice. This is because even for moderate $n$, the number of configurations $N=c^n$ is large and thus the number of random samples required to encounter any given one grows quickly in general.

\section{Conclusion}\label{sec:Conclusion}
We have introduced a stable and efficient flow-matching approach on the statistical assignment manifold which enables to learn complex joint distributions of many discrete random variables. 
While some limitations related to likelihood estimation remain to be adressed in future work, our approach shows clear promise as an alternative to score- or likelihood-based modelling for the challenging setting of discrete data.
This promise is founded on principled information-geometric modelling of discrete distributions.

%\begin{contributions} % will be removed in pdf for initial submission 
%					  % (without ‘accepted’ option in \documentclass)
%                      % so you can already fill it to test with the
%                      % ‘accepted’ class option
%    Briefly list author contributions. 
%    This is a nice way of making clear who did what and to give proper credit.
%    This section is optional.
%
%    H.~Q.~Bovik conceived the idea and wrote the paper.
%    Coauthor One created the code.
%    Coauthor Two created the figures.
%\end{contributions}

\begin{acknowledgements} % will be removed in pdf for initial submission,
						 % (without ‘accepted’ option in \documentclass)
                         % so you can already fill it to test with the
                         % ‘accepted’ class option
This work is funded by the Deutsche Forschungsgemeinschaft (DFG), grant SCHN 457/17-1, within the priority programme SPP 2298: ``Theoretical Foundations of Deep Learning''.
This work is funded by the Deutsche Forschungsgemeinschaft (DFG) under Germany's Excellence Strategy EXC-2181/1 - 390900948 (the Heidelberg STRUCTURES Excellence Cluster).
\end{acknowledgements}

%\clearpage
% References
\newpage
\bibliography{bb,cs}

\newpage

\onecolumn

\title{Generative Modeling of Discrete Joint Distributions by \\ E-Geodesic Flow Matching on Assignment Manifolds \\(Appendix)}
\maketitle

\appendix
\setcounter{page}{11}
\section{Implementation Details}\label{app:experiment_details}
% !TEX root =  ../AF_Generative-Discrete_uai2024.tex

For the Cityscapes experiment, we employ the UNet architecture of \cite{Dhariwal2021diffusion} with 
\emph{attention$\_$resolutions} (32, 16, 8), \emph{channel$\_$mult} (1,1,2,3,4), 4 attention heads, 3 blocks and 64 channels.
We trained for 250 epochs using Adam with learning rate 0.0001 and cosine annealing scheduler.

\begin{figure}[h]
	\centering
	\begin{subfigure}{1.\textwidth}
		\includegraphics[width=\textwidth]{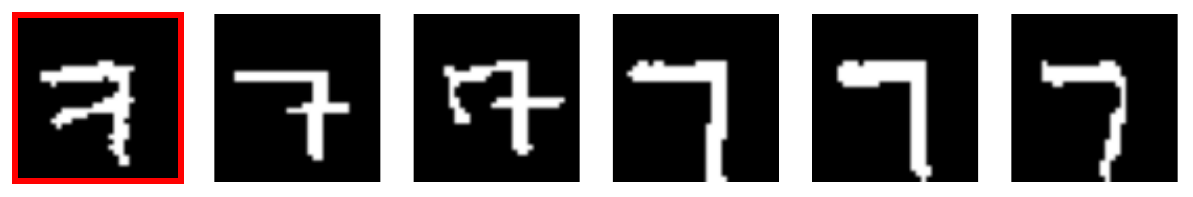}
		%\caption{SubFig 1}
	\end{subfigure}
%%%%%%%%%%%%%%
	\begin{subfigure}{1.\textwidth}
		\includegraphics[width=\textwidth]{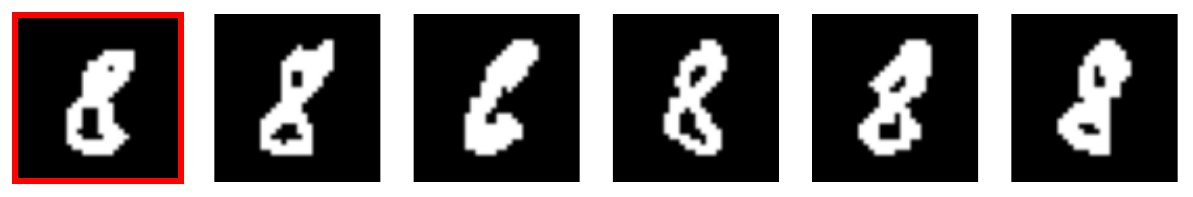}
	\end{subfigure}
%%%%%%%%%%%%%%
	\begin{subfigure}{1.\textwidth}
		\includegraphics[width=\textwidth]{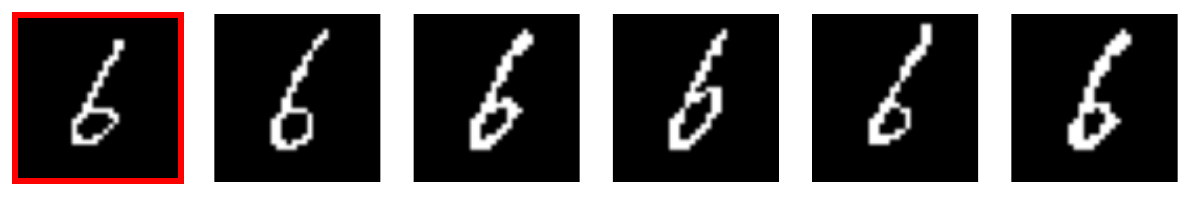}
	\end{subfigure}
%%%%%%%%%%%%%%
	\begin{subfigure}{1.\textwidth}
		\includegraphics[width=\textwidth]{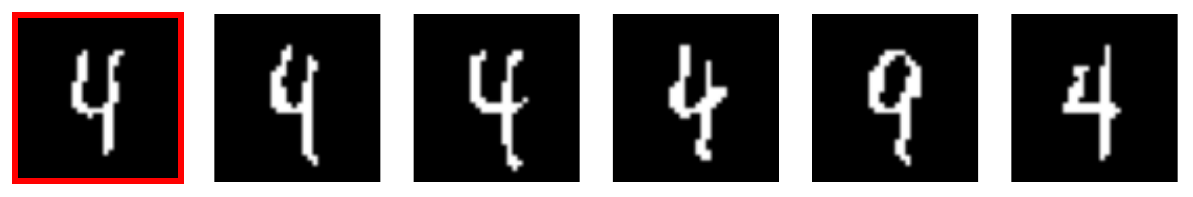}
	\end{subfigure}
	\caption{Illustration of handwritten digit samples trained from the MNIST dataset. The leftmost column, highlighted with a red frame, displays random samples generated by our model through integration Eq. \eqref{eq:def-AF} with random initializations. The remaining plots depict the five closest training data to our sample based on pixel-wise distances.} \label{fig:MNIST_similarity}
\end{figure}

For MNIST, we use the same architecture with \emph{attention$\_$resolutions} (16), \emph{channel$\_$mult} (1,2,2,2), 4 attention heads, 2 blocks and 32 channels. We trained for 100 epochs using Adam with learning rate 0.0005 and cosine annealing scheduler..
We pad the original $28\times 28$ images with zeros to size $32\times 32$ to be compatible with spatial downsampling employed by the UNet architecture. As Figure \ref{fig:MNIST_similarity} shows, our model  does not simply memorize the training data.

For the simple distributions in Figure~\ref{fig:simple_distr}, we employ a neural network composed of batch normalization, dense layers and ReLU activation. The sequence of hidden dimensions for the mixture of Gaussian and Pinwheel distributions is (256, 256). For the coupled binary variables, we use a linear function $F_\theta$, with no batch normalization or bias.
We trained for 2k steps with batch size 512 using Adam with learning rate 0.0005.

In all experiments, the smoothing constant $\epsilon$ of \eqref{eq:smoothed_corner} is set to $0.01$.

Rather than the original $c=33$ classes, we only use the $c=8$ class categories specified in \emph{torchvision}. The same subsampling of classes was used in the related work \cite{Hoogeboom:2021aa}. They additionally perform spatial subsampling to $32\times 64$. Instead, we subsample the spatial dimensions (\emph{NEAREST} interpolation) to $128\times 256$.

All experiments in this paper were run on one of two desktop graphics cards (1x NVIDIA RTX2080ti, 1x NVIDIA RTX2060super), requiring less than 100 compute hours in total.

\begin{figure}[h]
	\centering
	\begin{subfigure}{1.\textwidth}
		\includegraphics[width=\textwidth]{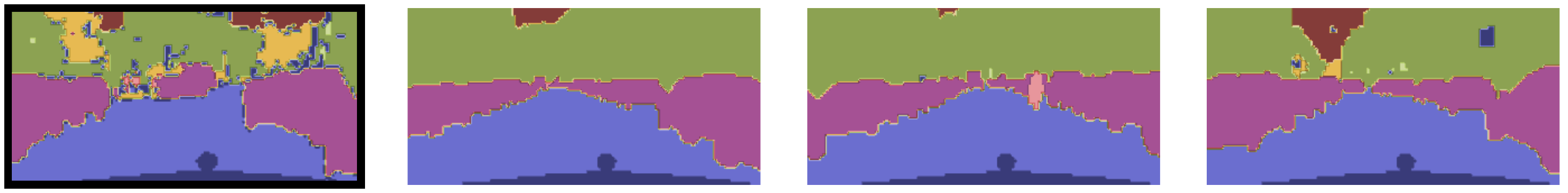}
		%\caption{SubFig 1}
	\end{subfigure}
%%%%%%%%%%%%%%
	\begin{subfigure}{1.\textwidth}
		\includegraphics[width=\textwidth]{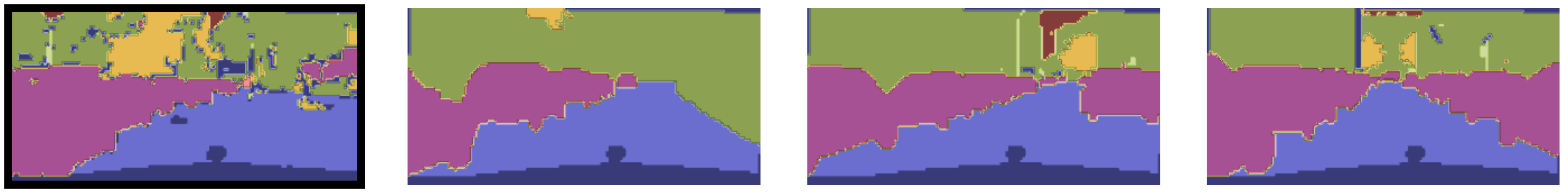}
	\end{subfigure}
%%%%%%%%%%%%%%
	\begin{subfigure}{1.\textwidth}
		\includegraphics[width=\textwidth]{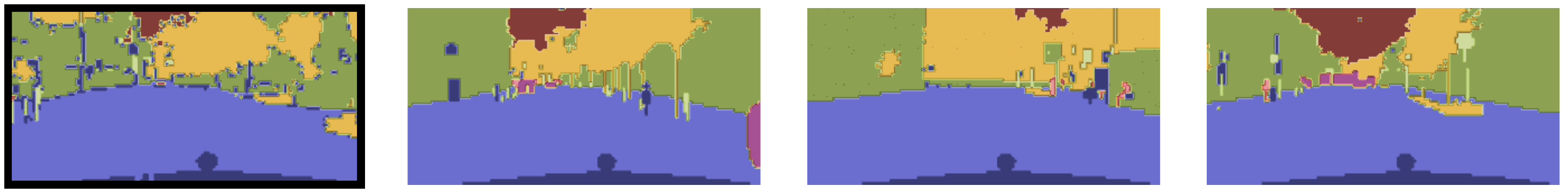}
	\end{subfigure}
%%%%%%%%%%%%%%
	\begin{subfigure}{1.\textwidth}
		\includegraphics[width=\textwidth]{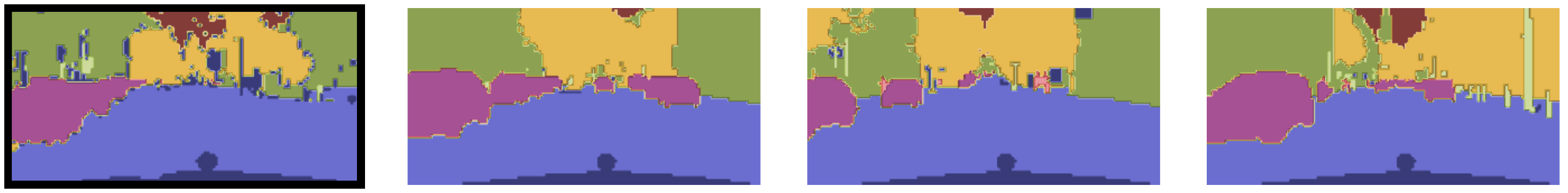}
	\end{subfigure}
	\caption{Illustration of Cityscapes segmentation samples drawn from our model. The leftmost column, highlighted with a black frame, displays random samples generated by our model through integration Eq. \eqref{eq:def-AF} with random initializations. The remaining plots depict the five closest training data to sample based on pixel-wise distance.}
 \label{fig:cityscapes_similarity}
\end{figure}

\section{Importance Sampling}\label{sec:importance_sampling_computations}
% !TEX root =  ../AF_Generative-Discrete_uai2024.tex

Let $Q\in\R^{c\times (c-1)}$ be an orthonormal basis of the linear subspace $T_0\mc{S}_c\subseteq \R^{c}$.
Independently for the tangent space of every individual simplex with index $i\in [n]$, the chosen proposal distribution is the normal distribution
\begin{equation}\label{eq:prop_distr_coord}
    \rho_i = Q_\sharp \mc{N}\big(Q^\top (\wt q_\alpha)_i, \sigma^2\II_{c-1}\big)
\end{equation}
on $T_0\mc{S}_c$, centered and supported on a disk with radius $r>0$,
\begin{equation}
    \wt r_\alpha = \{u_i\in T_0\mc{S}_c\colon \|u_i-(\wt q_\alpha)_i\|_2 \leq r\}.
\end{equation}
For any $u_i\in T_0\mc{S}_c$, it holds $QQ^\top u_i = u_i$ and we have
\begin{subequations}
\begin{align}
\|u_i-(\wt q_\alpha)_i\|_2^2 
    &= \la u_i-(\wt q_\alpha)_i, u_i-(\wt q_\alpha)_i\ra\\
    &= \la QQ^\top(u_i-(\wt q_\alpha)_i), u_i-(\wt q_\alpha)_i\ra\\
    &= \la Q^\top(u_i-(\wt q_\alpha)_i), Q^\top(u_i-(\wt q_\alpha)_i)\ra\\
    &= \|Q^\top u_i-Q^\top(\wt q_\alpha)_i)\|_2^2.
\end{align}
\end{subequations}
Thus, $u_i\in \wt r_\alpha$ exactly if the coordinates $Q^\top u_i$ lie in the ball
\begin{equation}
    \wh r_\alpha = \{x\in \R^{c-1}\colon \|x-Q^\top (\wt q_\alpha)_i\|_2 \leq r\}.
\end{equation}
Since the proposal distribution \eqref{eq:prop_distr_coord} is a normal distribution with variance $\sigma^2$ centered and supported on $\wh r_\alpha$, we need the probability of $\wh r_\alpha$ under a normal distribution with full support centered on it, as normalization constant of $\rho_i$.
By first shifting the mean, this can be computed as the probability of the sphere $\{x\in \R^{c-1}\colon \|x\|_2\leq r\}$. Let $X$ be a standard normal random variable on $\R^{c-1}$, then the sought probability is
\begin{equation}\label{eq:ball_mass_chisq}
    \PP(\|\sigma X\|_2^2 \leq r^2) = \PP\Big(\|X\|_2^2\leq \frac{r^2}{\sigma^2}\Big).
\end{equation}
Since $X$ has normal distribution, $\|X\|_2^2$ has $\chi^2$-distribution and \eqref{eq:ball_mass_chisq} can be computed by evaluating the cumulative distribution function of $\chi^2$ with $c-1$ degrees of freedom.
In practice, we set a probability mass of 0.8 from the outset and then choose $\sigma^2$ by inverting \eqref{eq:ball_mass_chisq}.
A simple geometric argument shows that the largest radius $r$ which satisfies the requirements is
\begin{equation}
    r = \|\wt q_\alpha\|_2 \sqrt{\frac{c}{2(c-1)}}.
\end{equation}

%\section{Additional simulation results}\label{app:additional_simulations}
%\input{TexInput/Appendix_B}

%\section{Proofs}

\end{document}